%% file: main.tex
\documentclass{article}

\usepackage{microtype}
\usepackage{graphicx}
\usepackage{subfigure}
\usepackage{booktabs} 

\usepackage{multirow}
\usepackage{times}
\usepackage{latexsym}
\usepackage{amsmath}
\usepackage{amssymb}

\usepackage{stfloats}

\usepackage{hyperref}

\usepackage[accepted]{icml2021}

\newcommand\NLVR{$\text{NLVR}^2$}
\newcommand\ourst{VL-T5}
\newcommand\oursb{VL-BART}

\definecolor{demphcolor}{RGB}{144,144,144}
\newcommand{\demph}[1]{\textcolor{demphcolor}{#1}}

\icmltitlerunning{Unifying Vision-and-Language Tasks via Text Generation}

\begin{document}

\twocolumn[
\icmltitle{Unifying Vision-and-Language Tasks via Text Generation}

\icmlsetsymbol{equal}{*}
\begin{icmlauthorlist}
\icmlauthor{Jaemin Cho}{unc}
\icmlauthor{Jie Lei}{}
\icmlauthor{Hao Tan}{}
\icmlauthor{Mohit Bansal}{} \\
UNC Chapel Hill\\
\texttt{\{jmincho,jielei,haotan,mbansal\}@cs.unc.edu}\\

\end{icmlauthorlist}

\icmlaffiliation{unc}{UNC Chapel Hill}

\icmlcorrespondingauthor{Jaemin Cho}{jmincho@cs.unc.edu}

\icmlkeywords{Vision and Language, Machine Learning}

\vskip 0.3in
]

\printAffiliationsAndNotice{}  

\input{abstract}

\input{intro}

\input{related_works}

\input{architecture}

\input{pretrain}
\input{experiments}
\input{conclusion}

\bibliography{references}
\bibliographystyle{icml2021}

\clearpage

\appendix

\input{supplementary}

\end{document}

%% file: abstract.tex
\begin{abstract}
Existing methods for vision-and-language learning typically require designing task-specific architectures and objectives for each task.
For example, a multi-label answer classifier for visual question answering, a region scorer for referring expression comprehension, and a language decoder for image captioning, etc.
To alleviate these hassles, in this work, we propose a unified framework that learns different tasks in a single architecture with the same language modeling objective, i.e., multimodal conditional text generation, where our models learn to \emph{generate labels in text} based on the visual and textual inputs.
On 7 popular vision-and-language benchmarks, including visual question answering, referring expression comprehension, visual commonsense reasoning, most of which have been previously modeled as discriminative tasks, our generative approach (with a single unified architecture) reaches comparable performance to recent task-specific state-of-the-art vision-and-language models.
Moreover, our generative approach shows better generalization ability on questions that have rare answers.
Also, we show that our framework allows multi-task learning in a single architecture with a single set of parameters, achieving similar performance to separately optimized single-task models.
Our code is publicly available at: \url{https://github.com/j-min/VL-T5}

\end{abstract}

%% file: intro.tex
\section{Introduction}

Mirroring the success of the pretraining-finetuning paradigm with transformer language models \cite{Devlin2019}, recent vision-and-language transformers (\citet{Tan2019,Lu2019,Chen2020,Li2020a}, \textit{inter alia})
have also been adopted in a wide range of vision-and-language tasks. 
These models are firstly pretrained on large image-text corpus (e.g., COCO Caption~\cite{Chen2015c}), then finetuned on downstream tasks (e.g., visual question answering~\cite{Goyal2019} and referring expression comprehension~\cite{Mao2016}), which outperformed many previous non-pretraining-finetuning methods.

\input{fig_teaser}

For each pretraining or downstream task, existing vision-and-language transformers typically require designing task-specific, separately-parameterized architectures on top of the transformer encoder (e.g., multi-label sigmoid classifier for visual question answering, and softmax classifier for referring expression comprehension).
However, the reasoning skills required by these tasks overlap significantly.
Consider the example in Fig.~\ref{fig:teaser}.
Both answering the question ``What is the man jumping over?'' and grounding an image region corresponding to the phrase ``yellow fire hydrant'' require recognizing the object ``fire hydrant''. 
In addition, the labels for these tasks can be easily expressed in text.
For instance, we can assign a region id (e.g., ``\texttt{<vis\_3>}'', a special text token) to a specific region in the image, and then the referring expression comprehension task can be expressed as generating the correct region id. 
For visual question answering, the labels are already in text, although existing approaches~\cite{Anderson2018,Tan2019,Chen2020} tackle the task as learning a multi-label classifier over a fixed set of frequent answers (See Fig.~\ref{fig:comparison}).

Hence, in order to alleviate these hassles of designing task-specific architectures, we propose a unified framework for vision-and-language learning via \emph{generating labels in text}.
Specifically, we extend off-the-shelf pretrained language models T5 \cite{Raffel2019} and BART \cite{Lewis2020} with visual understanding ability, named `VL-T5' and `VL-BART'.
In contrast to existing methods that train different architectures for each pretraining and downstream task, our models tackle all tasks with the same language modeling head.
\textit{To learn a new task, we can simply rewrite its input and output in text}, without the need of adding extra parameters or designing new architectures and objectives.
In addition, \textit{we can leverage the text generation ability of pretrained language models when making predictions}.
This is especially helpful when we answer open-ended questions that require non-trivial answers, where discriminative methods can only answer from a predefined set of frequent candidates, while our models can generate open-ended natural language answers.

To evaluate the effectiveness of our generative modeling approach, we compare our models against recent vision-and-language transformers on a diverse set of 7 downstream benchmarks, including visual question answering on VQA~\cite{Goyal2019} and GQA~\cite{Hudson2019}, referring expression comprehension on RefCOCOg~\cite{Mao2016}, natural language visual reasoning on \NLVR{}\cite{Suhr2019}, visual commonsense reasoning on VCR~\cite{Zellers2019a}, image captioning on COCO Caption~\cite{Chen2015c}, and multimodal machine translation on Multi30K~\cite{Elliott2016}.
Our unified generative method reaches comparable performance to recent state-of-the-art vision-and-language pretraining methods.
This is especially interesting because we use the same unified language modeling architecture with the same maximum likelihood estimation (MLE) objective for all the tasks, while existing approaches use task-specific architectures and objective functions.
In addition, we found that our generative models have better generalization ability compared to the discriminative versions in the rare-answer scenario on visual question answering, when ground truth answers for given questions are rarely seen during training. 
Finally, we also experiment with our unified framework under the multi-task learning setup on all 7 downstream tasks.
With a single architecture and a single set of weights, our model achieves similar performance to separately optimized single-task models.

%% file: fig_teaser.tex
\begin{figure}[t]
\vskip 0.1in
\begin{center}
\includegraphics[
                width=0.9\columnwidth,
                ]{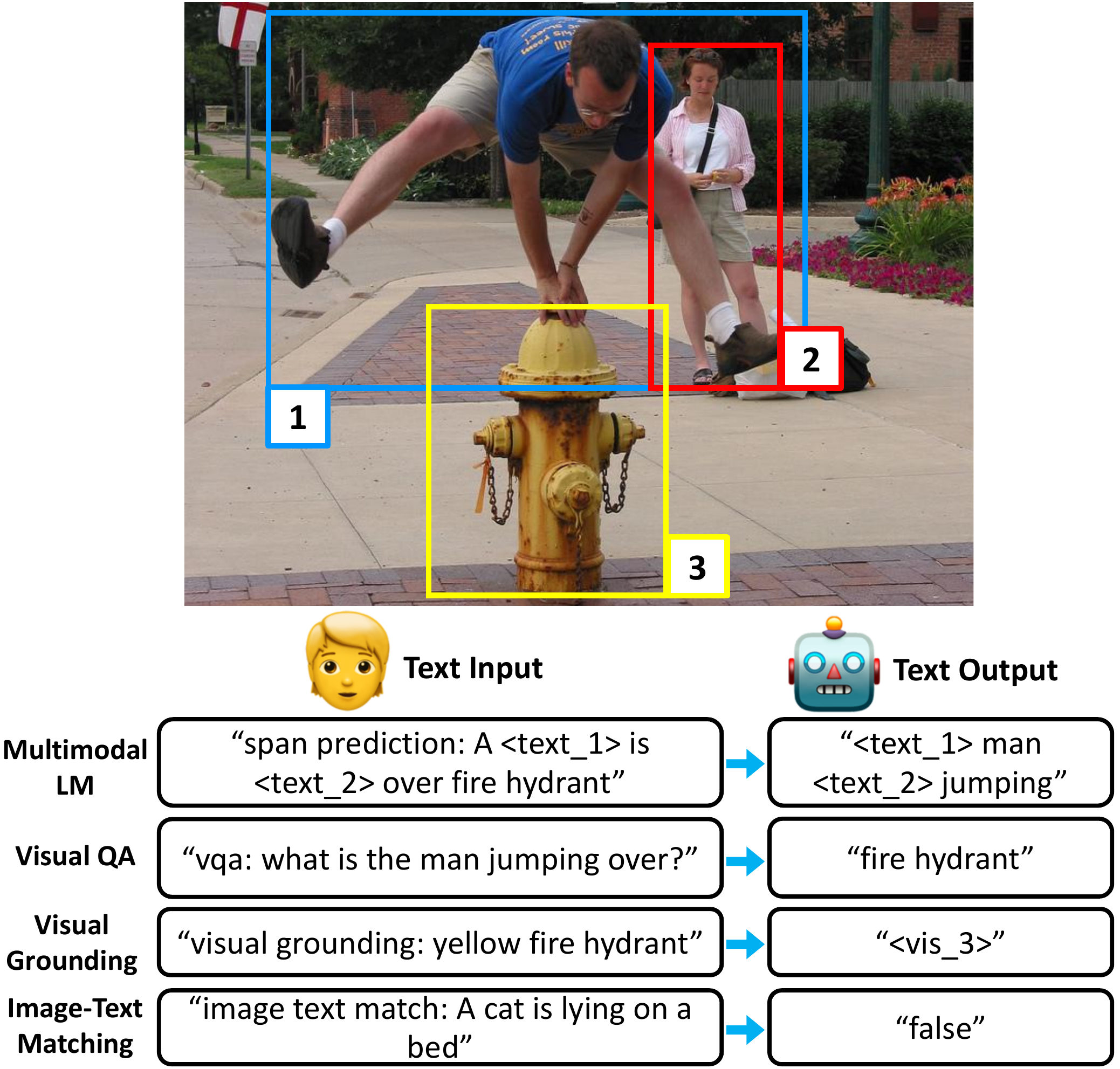}
                \vspace{-8pt}
\caption{
Our unified framework for learning vision-and-language tasks. 
While existing methods require designing task-specific architectures for different tasks, our framework unifies them together as generating text labels conditioned on multimodal inputs.
}
\end{center}
\label{fig:teaser}
\vspace{-10pt}
\end{figure}

%% file: related_works.tex
\section{Related Works}
\noindent\textbf{Vision-and-Language pretraining:} Large-scale language pretraining with transformers~\cite{Vaswani2017, Devlin2019,liu2019roberta,lan2019albert,clark2020electra,yang2019xlnet,Raffel2019} have achieved remarkable success for many natural language understanding tasks~\cite{rajpurkar2016squad,zellers2018swag,wang2018glue,williams2017broad}.
Following this success, image+text pretraining models~\cite{Lu2019,Tan2019,Chen2020,Huang2020,Li2020a,Cho2020,Radford2021,Zhang2021} and video+text pretraining models~\cite{Sun2019d,Sun2019c,Li2020b,zhu2020actbert,miech2020end} have also shown to perform better than previous non-pretraining approaches~\cite{Yu2018,Anderson2018,Kim2018,yu2018joint} in a wide range of discriminative~\cite{Goyal2019,Hudson2019,lei2018tvqa,Mao2016,xu2016msr,zhou2017towards} and generative tasks~\cite{Chen2015c,xu2016msr,zhou2017towards}.
In this work, we focus on image+text tasks.
While existing image+text models mostly use task-specific architectures and objectives, we seek to design a unified framework across different tasks.

\input{fig_architecture}

\noindent\textbf{Unified frameworks:} One line of work focus on solving natural language processing tasks in a unified format, such as question answering \cite{Mccann2018}, span prediction \cite{Keskar2019}, or text generation \cite{Raffel2019, Brown2020, Khashabi2020}.
These unified frameworks provide efficient knowledge sharing among different tasks and make it easy to leverage pretrained language models.
In relation to these works, we propose to unify previously separately modeled vision-and-language tasks in a single unified format, via text generation, conditioned on multimodal inputs from the image and the textual context.

%% file: fig_architecture.tex
\begin{figure*}[ht]
\begin{center}
\includegraphics[
                width=0.99\textwidth,
                 ]{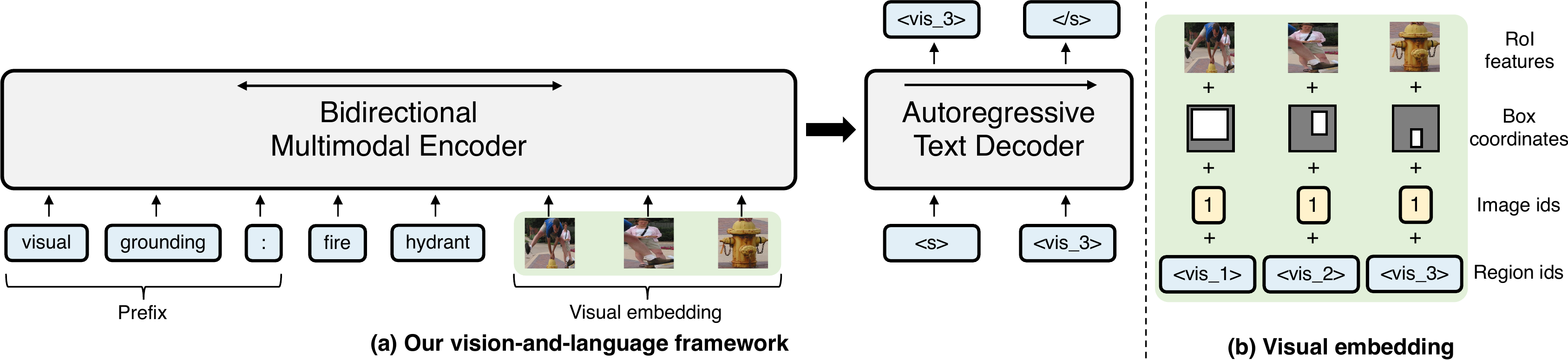}
                 \vspace{-8pt}
\caption{
An illustration of our \ourst{} and \oursb{} architectures for visual grounding task.
Instead of task-specific architectures, our models use text prefixes to adapt to different tasks.
The green block in (a) refers to visual embeddings.
(b) shows the components of visual embedding.
Note that we reuse the text embeddings of visual sentinel tokens (ex. \texttt{<vis\_3>}) as region id embeddings, which allows our models to tackle many discriminative vision-language tasks as text generation, including visual grounding.
}
\label{fig:architecture}
\end{center}
\vspace{-15pt}
\end{figure*}

%% file: architecture.tex
\section{Model}
\label{sec:model}
We propose a new framework that unifies vision-and-language problems as multimodal conditional text generation.
We introduce \ourst{} and \oursb{} based on two pretrained transformer language models: T5$_{Base}$ \cite{Raffel2019} and BART$_{Base}$ \cite{Lewis2020}.
Specifically, we extend their text encoders to multimodal encoders by incorporating image region embeddings as additional input.
The overall architecture of our framework is shown in Fig.~\ref{fig:architecture}.
Since the architecture differences between \ourst{} and \oursb{} are minor, we use \ourst{} as an example to illustrate our framework in details in the rest of this section.

\subsection{Visual Embeddings}
\label{sec:visual_embeddings}
We represent an input image $v$ with $n{=}36$ object regions
from a Faster R-CNN \cite{Ren2015} trained on Visual Genome \cite{Krishna2016} for object and attribute classification~\cite{Anderson2018}. 
As shown in Fig.~\ref{fig:architecture} (\textit{b}), each image region is encoded as a sum of four types of features:
($i$) RoI (Region of Interest) object features;
($ii$) RoI bounding box coordinates;
($iii$) image ids $\in \{1, 2\}$;
and 
($iv$) region ids $\in \{1, \dots, n\}$.
RoI features and bounding box coordinates are encoded with a linear layer, while image ids and region ids are encoded with learned embeddings~\cite{Devlin2019}.
Image ids are used to discriminate regions from different images, and is used when multiple images are given to the model (i.e., in \NLVR{}~\cite{Suhr2019}, models take two input images).
The final visual embeddings are denoted as $e^v {=} \{e^v_1, \dots, e^v_{n}\}$.

\subsection{Text Embeddings}
\label{sec:text_embeddings}
Instead of designing task-specific architectures, we add different prefixes to the original input text to adapt to different tasks, as shown in Table.~\ref{table:io-format}\footnote{Note that since we use simple prefixes (e.g., ``\texttt{vqa:}'' for VQA task), it is likely that engineering in text prompts \cite{Gao2020} would improve the accuracy of our methods. As this is not the focus of this paper, we leave it as future works.}.
This augmented input text $x$ is then tokenized as $\{x_1, \dots, x_{|x|}\}$ and encoded as learned embedding $e^x = \{e^x_1, \dots, e^x_{|x|}\}$.
The embedding parameters are shared by the encoder, decoder, and language modeling head \cite{Press2017}.
Since the attention layers are permutation-invariant, 
BART learns positional embeddings \cite{Vaswani2017, Devlin2019} for absolute token positions and adds them to the token embeddings.
In contrast, T5 adds relative position bias to 
each self-attention layer \cite{Shaw2018}. 
Our models follow the positional embedding configurations of their text backbone models.

In addition to the original vocabulary of T5 and BART, we introduce visual sentinel tokens \{\texttt{<vis\_1>}, $\dots$, \texttt{<vis\_n>}\}, which corresponds to image regions.
As illustrated in Fig.~\ref{fig:architecture}, we use the text embeddings of visual sentinel tokens as region id embeddings in Sec.~\ref{sec:visual_embeddings}.
The embedding sharing enables our model to build the correspondence among query text, label text, and objects, which are useful in the grounding tasks (e.g., visual grounding and grounded captioning pretraining tasks in Sec.~\ref{sec:pretraining}, referring expression comprehension in Sec.~\ref{sec:refcoco}).

\subsection{Encoder-Decoder Architecture}

We use transformer encoder-decoder architecture \cite{Vaswani2017} to encode visual and text inputs and generate label text.
Our bidirectional multimodal encoder is a stack of $m$ transformer blocks, consisting of a self-attention layer and a fully-connected layer with residual connections.
Our decoder is another stack of $m$ transformer blocks similar to the multimodal encoder, where each block has an additional cross-attention layer.
As shown in Fig.~\ref{fig:architecture} (\textit{a}), the encoder takes the concatenation of text and visual embeddings as input and outputs their contextualized joint representations $h = \{h^x_{1}, \dots, h^x_{|x|}, h^v_{1}, \dots, h^v_{n}\} = \mathrm{Enc}(e^x, e^v)$.
Then the decoder iteratively attends to previously generated tokens $y_{<j}$ (via self-attention) and the encoder outputs $h$ (via cross-attention), then predicts the probability of future text tokens
$P_{\theta}(y_{j} |y_{< j}, x, v) = \text{Dec}(y_{<j}, h)$.
We suggest readers to check \citet{Raffel2019, Lewis2020} for more details of our backbone models.
For both pretraining (Sec.~\ref{sec:pretraining}) and downstream tasks (Sec.~\ref{sec:downstream}), 
we train our model parameters $\theta$ by minimizing the negative log-likelihood of label text $y$ tokens given input text $x$ and image $v$:
\begin{equation}
\label{eq:loss_gen}
\mathcal{L}_\theta^{\textsc{gen}} = - \sum_{j=1}^{|y|}\log P_{\theta}(y_{j} |y_{< j}, x, v)
\vspace{-8pt}
\end{equation}

\subsection{Task-Specific Methods \textit{vs.} Our Unified Framework}
\label{subsec:comparison}

We compare our unified framework with existing vision-and-language transformers on two popular tasks: visual question answering~\cite{Goyal2019} and referring expression comprehension~\cite{Mao2016}.

\noindent\textbf{Visual question answering}
requires a model to answer a question to a given context image.
As shown in Fig.\ref{fig:comparison} (\textit{a}), existing methods \cite{Tan2019, Lu2019, Chen2020} typically introduce a multi-layer perceptron (MLP) multi-label classifier head on top of $h^x_{\texttt{[CLS]}}$, which is trained together with the transformer backbone through a binary cross-entropy loss, and weighted with VQA score \cite{Goyal2019}\footnote{$\text{score}(a,x,v) {=} \min(\text{\small(\#humans that gave } \text{answer a)} * 0.3,1)$}: $\mathcal{L}^{\textsc{vqa}}_{\theta} {=} - \sum_{k=1}^{K} \text{score}(a^k,x,v)\log P^\textsc{vqa}_{\theta}(\text{correct}|a^k,x,v)$.

\input{fig_comparison_topvsbottom}

\noindent\textbf{Referring expression comprehension}
requires models to localize a target region in an image that is described by a given referring expression.
Previous methods tackle this task as multi-class \cite{Chen2020} or binary \cite{Lu2019} classification over image regions.
For example, UNITER~\cite{Chen2020} introduces an MLP region scoring head on top of the output representations of regions, as shown in Fig.~\ref{fig:comparison}(\textit{b}).
This region scoring head is jointly trained with the encoder by minimizing negative log-likelihood of target region $r^*$: $\mathcal{L}^{\textsc{ref}}_{\theta} = - \log P^{\textsc{ref}}_{\theta}(r^{*}|x,v)$.

In contrast to existing methods that develop task-specific architectures and objectives (e.g., equations above), our unified framework is free from extra model designs for new tasks.
As shown in Fig.~\ref{fig:comparison} (\textit{c,d}) and Table~\ref{table:io-format}, we formulate the task labels to corresponding text, and we learn these different tasks by predicting label text with the same language modeling objective (Eq.~\ref{eq:loss_gen}).

%% file: fig_comparison_topvsbottom.tex
\begin{figure}[t]
\begin{center}
\includegraphics[
                width=0.99\columnwidth, ]{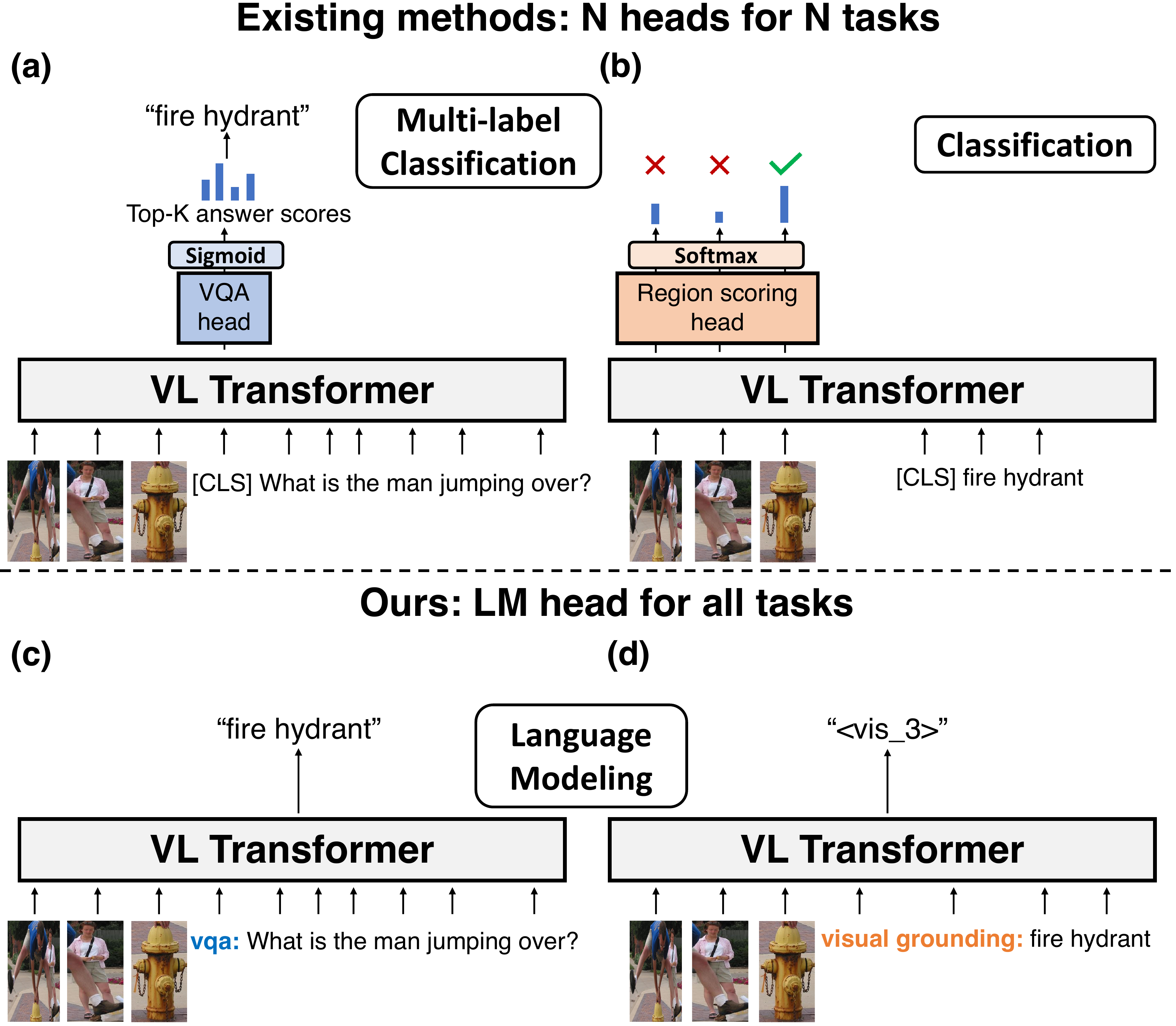}
\vspace{-6pt}
\caption{
Comparison between existing methods and our framework on visual question answering and referring expression comprehension (visual grounding) tasks.
While existing methods use task-specific architectures and objectives, our models use the same language modeling architecture and maximum likelihood estimation on label text for all tasks.
}
\label{fig:comparison}
\end{center}
\vspace{-20pt}
\end{figure}

%% file: pretrain.tex
\section{Pretraining}
\label{sec:pretraining}

\input{table_task_io}

In this section, we describe how we pretrain our \ourst{} and \oursb{} models (Sec.~\ref{sec:model}). We start with the details of the pretraining data and illustrate how we formulate diverse vision-and-language pretraining tasks as multimodal conditional text generation.

\subsection{Pretraining Data}
We aggregate pretraining data from MS COCO~\cite{Lin2014, Chen2015c} and Visual Genome (VG; \citet{Krishna2016}) images\footnote{Existing vision-and-language transformers are trained with different datasets and computational budgets, thus their results may not be directly comparable to each other. We show the number of their pretraining images in Table~\ref{table:test-set}.}.
The captioning data from these two datasets are used in the multimodal language modeling task.
The COCO captions are also used in the image-text matching task to learn cross-modal alignment.
Besides the captions, we also use three visual question answering datasets (VQA v2.0~\cite{Goyal2019}, GQA balanced version~\cite{Hudson2019}, and Visual7W~\cite{Zhu2016}) as in \citet{Tan2019}, but only used them for the visual question answering task.
Details of these pretraining tasks are in Sec.~\ref{sec:pretraining_task}.
Overall, our pretraining dataset contains 9.18M image-text pairs on 180K distinct images.
We show more details of the pretraining data in appendix.

\subsection{Pretraining Tasks}
\label{sec:pretraining_task}
We pretrain our models under a multi-task setup with diverse pretraining tasks,
including multimodal language modeling, visual question answering, image-text matching, visual grounding, and grounded captioning. 
Table~\ref{table:io-format} shows input and output examples of our pretraining tasks.
The training data for each of these tasks are summarized in
appendix.
In the rest of this section, we explain these tasks in detail.

\noindent\textbf{Multimodal language modeling:}
We follow \citet{Raffel2019} and \citet{Lewis2020} to construct the language modeling pretraining task. 
For \ourst{}, we mask 15\% of input text tokens and replace contiguous text span with sentinel tokens (e.g., \texttt{<text\_1>}).
For \oursb{}, we mask 30\% of input text tokens with \texttt{<mask>} tokens. 
Then we predict the masked text. See Table~\ref{table:io-format} for examples.

\noindent\textbf{Visual question answering:}
We include visual question answering in our pretraining tasks as in \citet{Tan2019}.
While previous methods \cite{Tan2019, Lu2019, Chen2020} tackle the task as classification over predefined answer candidates (illustrated in Fig.~\ref{fig:comparison}), we directly generate answers in their original text format.

\noindent\textbf{Image-text matching:}
In this task, the model needs to verify whether a text corresponds to an image.
We consider an image and its captions\footnote{We only use captions from COCO for this task, since many short captions from VG and visual questions are
nondistinctive descriptions of an image (e.g., `what is in the image?').} as positive pairs. 
With a probability of 50\%, we randomly sample another training image's caption to create a negative pair.
The model then predicts the correspondence
with ``true'' or ``false'' as shown in Table~\ref{table:io-format}.

\noindent\textbf{Visual grounding:}
We develop an object-text matching task to endow the model with grounding ability, which is required in several tasks (e.g., referring expression comprehension and VCR).
We give the model a region description and let it predict the id of the related object region.
With the help of the visual sentinel token (e.g., \texttt{<vis\_3>} in Table~\ref{table:io-format}), this task fits naturally into our text generation objective.
We make the region descriptions from the predictions of the object detector that we use for visual embeddings (see Sec. \ref{sec:visual_embeddings}).
Concretely, we sample an object region out of $n$ region predictions.
Then we concatenate its object name and attribute (e.g., attribute: ``yellow" + object: ``fire hydrant" $\rightarrow{}$ ``yellow fire hydrant").
This approach does not need extra annotation and could be extended to images without dense annotations (e.g., COCO images).

\noindent\textbf{Grounded captioning:}
To teach the model with object-level information, we also use grounded captioning as an inverse task of visual grounding.
As shown in Table~\ref{table:io-format}, given a visual sentinel token (which indicates an image region) as text input, the model is asked to generate a corresponding textual description of the image region.

\subsection{Pretraining Implementation Details}
For both \ourst{} and \oursb{}, it takes 4 days for 30-epoch pretraining with mixed precision training \cite{Narang2018} on 4 RTX 2080 Ti GPUs.
We use batch size 320 and 600 for \ourst{} and \oursb{}, respectively. 
We use AdamW \cite{Loshchilov2019} with $(\beta^1, \beta^2)=(0.9, 0.999)$ and learning rate 1e-4 with 5\% linear warmup schedule.
Our code is based on PyTorch \cite{Paszke2017} and Huggingface Transformers \cite{Wolf2019}.

%% file: table_task_io.tex
\begin{table*}[!t]
\caption{
Input-output formats for pretraining (Sec.~\ref{sec:pretraining}) and downstream tasks (Sec.~\ref{sec:downstream}).
$^a$We use different prefixes (``vqa:'', ``gqa:'', ``visual7w:'') for questions from different datasets.
$^b$\NLVR{} takes two images as visual input, for brevity, we only show one here.
}
\label{table:io-format}
\vskip 0.15in
\centering
\resizebox{\textwidth}{!}{
\begin{tabular}{llll}
\toprule
Tasks & Input image & Input text & Target text  \\
\midrule
\bf Pretraning tasks (Sec.~\ref{sec:pretraining}) & \multirow{17}{*}{\raisebox{-.5\totalheight}{\includegraphics[width=.8\columnwidth]{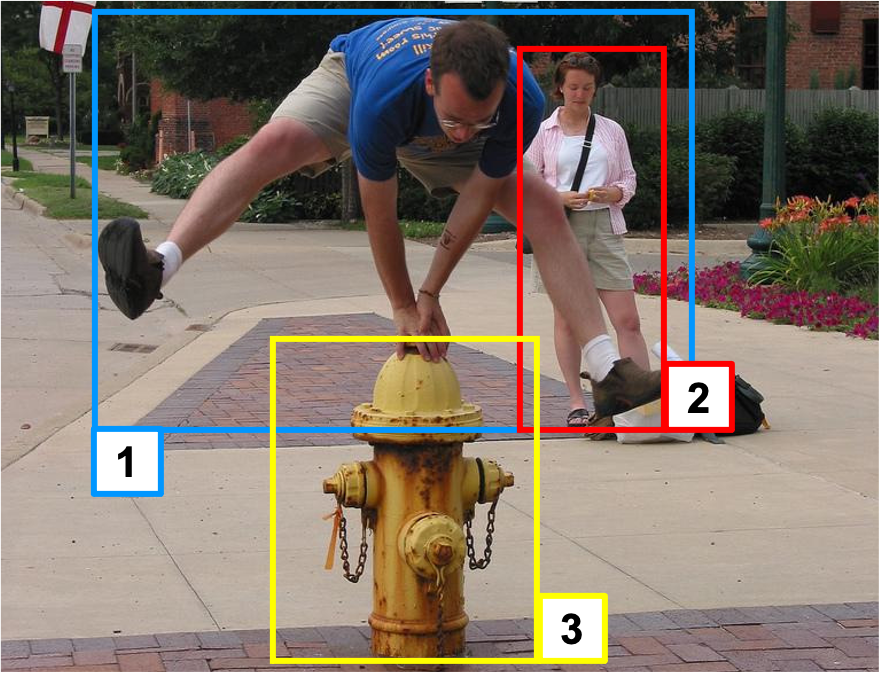}}} \\
Multimodal LM (\ourst{}) & & span prediction: A \texttt{<text\_1>} is \texttt{<text\_2>} over a fire hydrant. & \texttt{<text\_1>} man \texttt{<text\_2>} jumping \\
Multimodal LM (\oursb{}) & & denoise: A \texttt{<mask>} is \texttt{<mask>} over a fire hydrant. & A man is jumping over a fire hydrant \\ 
$^a$Visual question answering & & vqa: what is the color of the man's shirt? & blue \\
Image-text matching & & image text match: A man with blue shirt is jumping over fire hydrant. & true \\
Visual grounding & & visual grounding: yellow fire hydrant & \texttt{<vis\_3>} \\
Grounded captioning & & caption region: \texttt{<vis\_3>} & yellow fire hydrant \\
\cmidrule{1-1}\cmidrule{3-4}
\bf Downstream tasks (Sec.~\ref{sec:downstream}) \\
VQA & & vqa: \texttt{[Q]} & \texttt{[A]} \\
GQA & & gqa: \texttt{[Q]} & \texttt{[A]} \\
$^b$\NLVR{} & & nlvr: \texttt{[text]} & true/false \\
VCR Q$\rightarrow$A & & vcr qa: question \texttt{[Q]} answer: \texttt{[A]} & true/false \\
VCR QA$\rightarrow$R & & vcr qar: question \texttt{[Q]} answer: \texttt{[A]} rationale: \texttt{[R]} & true/false \\
RefCOCOg & & visual grounding: \texttt{[referring expression]} & \texttt{[region id]} \\
COCO captioning & & caption: & \texttt{[caption]} \\
COCO captioning (w/ object tags) & & caption with tags: \texttt{[Tag1 Tag2 ..]} & \texttt{[caption]} \\
Multi30K En-De translation & & translate English to German: \texttt{[English text]} & \texttt{[German text]} \\
\bottomrule
\vspace{-20pt}
\end{tabular}
}
\end{table*}

%% file: experiments.tex
\input{table_test}

\section{Downstream Tasks and Results}
\label{sec:downstream}

In this section, we compare our generative architectures \ourst{} and \oursb{} on a diverse set of 7 downstream tasks (details in Appendix) with existing vision-and-language pretrained transformers \cite{Tan2019, Lu2019, Chen2020, Zhou2020, Li2020a, Xia2020}.
As summarized in Table~\ref{table:test-set}, our unified generative approach (with the input-output format in Table~\ref{table:io-format}) shows performance close to the task-specific models, most of which are discriminative.
In the rest of this section, we provide detailed comparisons w.r.t. the baselines.

\subsection{Visual Question Answering: VQA and GQA}
The visual question answering task requires models to answer a question to a given context image.
Table \ref{table:test-set} compares our models \ourst{} and \oursb{} with existing methods on VQA \cite{Goyal2019} and GQA \cite{Hudson2019}. 
For both tasks, our models achieve comparable performance to existing approaches.

\input{table_vqa_oov}

\noindent\textbf{Generative \textit{vs.} Discriminative model:}
Modern approaches~\cite{Tan2019,Lu2019,Chen2020,Zhou2020,Li2020a} are discriminative models, where they tackle visual question answering tasks as multi-label classification over a predefined set of answer candidates. 
This strategy achieves strong performance but not generalizes to real-world open-ended scenarios.
To quantitatively compare the existing discriminative approaches and our generative approach, we break down VQA questions into in-domain and out-of-domain questions, in terms of whether the best answer $a^*$ for each question is included in the top-K ($K{=}3,129$) answer candidates $A^{topk}$.
After this split, the in-domain subset contains 24,722 questions, and the out-of-domain subset contains 1,558 questions.
Table~\ref{table:vqa_oov} shows the performance.
For discriminative baselines, we introduce a sigmoid MLP classifier on top of the decoder representation of \textit{start-of-sequence} token \texttt{<s>}, following LXMERT and UNITER.
Comparing models with the same backbone, we notice the generative models improve upon the discriminative baselines across all the subsets. 
This improvement is more significant on the out-of-domain subset, where the generative \ourst{} and \oursb{} achieve 6 and 6.2 points improvement over their discriminative counterparts, showing the effectiveness of using generative modeling.
Compared to the strong discriminative baseline UNITER$_{Base}$ (pretrained with 4M extra images), our generative models still show comparable overall performance while significantly outperform it on the out-of-domain subset (about 3 points).

\noindent\textbf{Dataset-specific prefixes:}
As shown in recent works \cite{Gao2020, autoprompt:emnlp20, li2021prefixtuning, Radford2021}, different text prompts could result in different finetuning results.
We thus experiment with a single prefix `vqa' for both VQA and GQA in \ourst{} pretraining/finetuning.
Interestingly, we found slight performance increases from the original dataset-specific prefix:
VQA Karpathy-test ($67.9\rightarrow69.3$); 
GQA test-dev ($60.0\rightarrow60.2$).
This shows that a single model can successfully handle multiple VQA tasks without dataset-specific prefixes (similar results were observed in text QA \cite{Khashabi2020}).

\input{table_nlvr}

\subsection{Natural Language Visual Reasoning: \NLVR{}}
The task of \NLVR{}~\cite{Suhr2019} is to determine whether a natural language statement is true about two images.
To apply our model to this task, we concatenate region features from the two images and use different image id embeddings to disambiguate the regions from the two images. 
Then our model learns to generate text labels ``true'' and ``false''.
This is similar to the \textit{Triplet} setting described in UNITER \cite{Chen2020}.
In Fig.~\ref{fig:nlvr_settings}, we illustrate three common encoding settings for \NLVR{}.

\input{fig_nlvr_settings}

Table~\ref{table:nlvr} shows the model results on \NLVR{} under different encoding settings:
($i$) \textit{Triplet}: joint encoding of image pairs and text;
($i$) \textit{Pair}: the concatenation of individual embedding of each image-text pair;
($iii$) \textit{Pair-biattn}: bidirectional attention added to \textit{Pair}.
UNITER shows that one can improve performance with a more complex encoding setting, i.e., \textit{Pair-biattn} achieves better performance than \textit{Pair}, which is again better than the simplest \textit{Triplet}.
Note that both the \textit{Pair} and the \textit{Pair-biattn} settings approximately double the computational cost compared to that of the \textit{Triplet} setting.
While there's the gap between our models and baselines in \textit{Pair} and \textit{Pair-biattn} setting, \ourst{} shows comparable performance to UNITER in \textit{Triplet} setting.

\subsection{Referring Expression Comprehension: RefCOCOg}
\label{sec:refcoco}
Referring expression comprehension requires a model to correctly localize an object described by a given phrase (e.g., `the car on the left').
In this work, we evaluate models on the RefCOCOg \cite{Mao2016} dataset.
Similar to the visual grounding pretraining task in Sec.~\ref{sec:pretraining}, we give our model a referring phrase and candidate region features from the image, the model then generates the visual sentinel token (e.g., \texttt{<vis\_1>}) of the region corresponding to the phrase.
Following previous works UNITER and MAttNet~\cite{Yu2018}, we use region detections from Mask R-CNN~\cite{He2017} as candidates and mark a selected region to be correct if its intersection over union (IoU) with the ground truth region is greater than 0.5.

Table~\ref{table:test-set} compares our models with discriminative baselines. 
With pretraining, \ourst{} significantly outperforms the strong modular model MAttNet, and achieves a reasonable performance compared to the UNITER model that has been pretrained on a much larger corpus.  
While our method did not achieve state-of-the-art performance, these results suggest that referring expression comprehension can be effectively formulated as a text-generation task, rather than previously~\cite{Yu2018,Chen2020} formulated classification task over a set of visual regions, allowing more flexible architecture design.
We hope our work would inspire future works in this direction.
We also observe that our experiments with \oursb{} on RefCOCOg diverges.
One reason might be the difference in positional encoding methods of T5 and BART. During training, BART adds learned absolute positional embedding to text token embedding, whereas T5 uses relative position biases in self-attention layers instead.
We hypothesize that \oursb{} found strong correspondence by memorizing the positions of each training object (we observe high training accuracy, but low validation accuracy).

\input{table_refcoco}

\input{table_vcr_full}

\subsection{Visual Commonsense Reasoning: VCR}

Visual Commonsense Reasoning (VCR) \cite{Zellers2019a} is a multiple-choice question answering task that requires commonsense reasoning beyond object or action recognition.
Each VCR question (Q) has 4 answers (A) and 4 rationales (R), and it can be decomposed into two multiple choice sub-tasks: question answering (Q$\rightarrow$A), and answer justification (QA$\rightarrow$R). 
The overall task (Q$\rightarrow$AR) requires a model to not only select the correct answer to the question, but also the correct rationale for choosing the answer.
Similar to \citet{Nogueira2020} that leverages language model for document ranking, we concatenate context (image+question) with each candidate choice, and let our models generate ``true'' for the correct choice and generate ``false'' otherwise, as shown in Table~\ref{table:io-format},
During inference, we use $\frac{P(true)}{P(true)+P(false)}$ to rank the choices and select the one with the highest score.

UNITER \cite{Chen2020} has shown that a second-stage in-domain pretraining (with the same pretraining objectives as generic-domain pretraining) on the VCR dataset would significant improve VCR task performance. 
This is likely due to the domain difference between VCR and the generic-domain pretraining corpus (e.g., COCO Captions), e.g., the input text (concatenation of multiple sentences: \texttt{[Q]+[A]+[R]}) in VCR is much longer than in generic-domain pretraining.
In Table~\ref{table:vcr_full}, we show the experiment results with second stage pretraining on VCR.
On VCR val split, comparing to the base models that do not pretrain, we find that both Stage 1 generic-domain pretraining and Stage 2 in-domain pretraining help improve the VCR task performance, which is consistent with the findings in UNITER.
On VCR test split, we notice that our best model \ourst{} achieves a comparable (slightly better) performance to UNITER, while significantly higher performance when compared to ViLBERT.

\input{table_caption}

\input{table_mmt}

\subsection{Image Captioning: COCO Caption}
We evaluate automatic caption generation performance on MS COCO Caption dataset \cite{Chen2015c}.
We use \textit{Karparthy split} \cite{Karpathy2015}, which re-splits train2014 and val2014 images \cite{Lin2014} into 113,287 / 5000 / 5000 for train / validation / test.
While some methods use reinforcement learning-based optimization on CIDEr, we only compare with methods using cross-entropy loss.
Note that image captioning is the only task in our experiments where textual context is not meaningful, which results in a notable difference in pretraining and finetuning w.r.t. the input format.
Inspired by Oscar \cite{Li2020b}, we also experiment with using object tags as additional text inputs during finetuning.
We use BLEU \cite{Papineni2002}, CIDEr \cite{Vedantam2015}, METEOR \cite{Banerjee2005}, SPICE \cite{Anderson2016} as evaluation metrics using COCOEvalCap\footnote{\url{https://github.com/tylin/coco-caption}}.

In Table~\ref{table:caption}, we compare our models with baselines in different settings: use of vision-and-language pretraining and use of object tag as additional text inputs.
With and without vision-and-language pretraining, our models show comparable performance to baselines.
Since the use of object tags requires significant extra computation, we only use it for finetuning. 
Using tags gives a comparable or slightly improved performance for both models, and the improvement is significant (2.5) in CIDEr for \oursb{}.
We expect object tag augmentation during pretraining like Oscar would further boost the performance of our models.

\subsection{Multimodal Machine Translation: Multi30K}
We evaluate multimodal machine translation performance on Multi30K dataset \cite{Elliott2016}, where a model translates English text to German text given context images.
We report BLEU score using SacreBLEU \cite{Post2018}\footnote{\url{https://github.com/mjpost/sacrebleu}}.
We compare our method with state-of-the-art transformer models: Multimodal self-attention (MSA) \cite{Yao2020}, MeMAD \cite{Gronroos2018}.
Table~\ref{table:mmt} shows that our T5-based models outperform the baselines that use strong data augmentation (e.g., back-translation) on all three test splits.
Our vision-and-language models improve the text-only backbones although we did not observe improvement with vision-and-language pretraining. This might be because the source text in Multi30K contains sufficient information for translation as discussed in \citet{Caglayan2019}

\subsection{Multi-Task Finetuning}
\noindent\textbf{Single-task \textit{vs.} Multi-task Finetuning:}
While our framework has unified the architecture for different downstream tasks, the parameters are separately optimized.
To see whether we can go further, we finetune a single \ourst{} model for 20 epochs, where it tackles 7 different tasks with the same set of weights.
At each finetuning step, we sample a mini-batch of examples from one of the 7 tasks in a round-robin fashion.
For a fair comparison, we use single-task baselines without augmentations
(e.g., no 2nd stage pretraining for VCR, no object tags for COCO Captioning).
Table~\ref{table:multitask-all} shows that our multi-task model achieves comparable performance to the separately optimized single-task models on all 7 tasks with a single set of parameters.

\input{table_multitask_all}

\noindent\textbf{Single shared head \textit{vs.} Task-specific heads:}
We also experiment with the multi-task finetuning setup of ViLBERT-MT \cite{Lu2020}, where a task-specific head is fine-tuned for each of 7 downstream tasks while sharing backbone. The head parameters are initialized from the pretrained LM head and separately updated during finetuning.
The 7 task-specific heads (7H) add $7 \times 32\mbox{K} (\mbox{vocab size}) \times 768 (\mbox{embedding size}) = 172\mbox{M}$ parameters, which is 80\% of original \ourst{}'s 220M parameters (P), resulting around 400M parameters in total.
Since the increased parameters make the training slow, we compare both models by 5th epoch checkpoints.
Table~\ref{table:multitask-FT-small} shows that \ourst{} with single shared head achieves almost equal performance with task-specific heads, while having much fewer total parameters.

\input{table_multitask_head}

%% file: table_test.tex
\begin{table*}[!t]
\caption{Single model performance on downstream tasks.
Note that the baseline models adopt task-specific objectives and architectures, whereas our models tackle all tasks, including discriminative tasks (e.g., RefCOCOg), as text generation with a single architecture and objective.
$\star$ See our discussion in Sec.\ref{sec:refcoco}.
}
\label{table:test-set}
\vskip 0.15in
\centering
\resizebox{0.98\textwidth}{!}{
\begin{tabular}{lcccccccc}

\toprule
\multirow{4}{*}{Method}& \multirow{4}{1.2cm}{\centering \# Pretrain Images} & \multicolumn{5}{c}{Discriminative tasks} & \multicolumn{2}{c}{Generative tasks} \\
\cmidrule(lr){3-7} \cmidrule(lr){8-9}
& & VQA & GQA & \NLVR{} & RefCOCOg & VCR Q$\rightarrow{}$AR & COCO Cap  & Multi30K En-De \\
& & test-std  & test-std & test-P & $\text{test}^d$ & test & Karpathy test & test 2018 \\
& & Acc & Acc & Acc & Acc & Acc & CIDEr & BLEU \\
\midrule
LXMERT  & 180K & 72.5 & 60.3 & 74.5 & - & - & - & - \\
ViLBERT & 3M & 70.9 & - & - & - & 54.8 & - & - \\
UNITER$_{Base}$  & 4M & 72.9 & - & 77.9 & 74.5 & 58.2 & - \\
Unified VLP & 3M & 70.7 & - & - & - & - & 117.7 & -\\
Oscar$_{Base}$ & 4M & 73.4 & 61.6 & 78.4 & - & - & 123.7 & - \\ 
XGPT & 3M & - & - & - & - & - & 120.1 & - \\
MeMAD & - & - & - & - & - & - & - & 38.5 \\
\midrule
\ourst{}  & 180K & 70.3 & 60.8 & 73.6 & 71.3 & 58.9 & 116.5 & 38.6 \\
\oursb{}  & 180K & 71.3 & 60.5 & 70.3 & 22.4$^\star$ & 48.9 & 116.6 & 28.1  \\
\bottomrule
\vspace{-20pt}
\end{tabular}
}
\end{table*}

%% file: table_vqa_oov.tex
\begin{table}[t]
\caption{VQA Karpathy-test split accuracy using generative and discriminative methods. 
We break down the questions into two subsets in terms of whether the best-scoring answer $a^*$ for each question is included in the top-K answer candidates $A^\mathit{topk}$. \textit{In-domain}: $a^* \in A^\mathit{topk}$, \textit{Out-of-domain}: $a^* \notin A^\mathit{topk}$.
}
\label{table:vqa_oov}
\vskip 0.15in
\centering
\resizebox{0.9\columnwidth}{!}{
\begin{tabular}{lccc}
\toprule
Method &  In-domain & Out-of-domain & Overall \\
\midrule
\multicolumn{2}{l}{\bf Discriminative} & & \\
UNITER$_{\text{Base}}$ & \textbf{74.4} & 10.0 & \textbf{70.5} \\
\ourst{}  & 70.2 & 7.1 & 66.4 \\
\oursb{}  & 69.4 & 7.0 & 65.7 \\
\midrule
\multicolumn{2}{l}{\bf Generative} & & \\
\ourst{} & 71.4 & 13.1 & 67.9 \\
\oursb{} & 72.1 & \textbf{13.2} & 68.6 \\
\bottomrule
\vspace{-20pt}
\end{tabular}
}
\end{table}

%% file: table_nlvr.tex
\begin{table}[t]
\caption{
\NLVR{} performance comparison under different encoding settings.
Note that \textit{Triplet} takes lower computational cost than \textit{Pair} and \textit{Pair-biattn}.
See also Fig.~\ref{fig:nlvr_settings}.
}
\label{table:nlvr}
\vskip 0.15in
\centering
\resizebox{0.8\columnwidth}{!}{
\begin{tabular}{llcc}
\toprule
Method & Setting & dev & test-P \\
\midrule
UNITER$_{Base}$ & Triplet & 73.0 & 73.9 \\
UNITER$_{Base}$ & Pair & 75.9 & 75.8 \\
UNITER$_{Base}$ & Pair-biattn & \textbf{77.2} & \textbf{77.9} \\
LXMERT & Pair & 74.9 & 74.5 \\
Oscar$_{Base}$ & Pair & 78.1 & 78.4 \\
\midrule
\ourst{} & Triplet & 74.6 & 73.6 \\
\oursb{} & Triplet & 71.7 &	70.3 \\
\bottomrule
\vspace{-20pt}
\end{tabular}
}
\end{table}

%% file: fig_nlvr_settings.tex
\begin{figure}[t]
\vskip 0.2in
\begin{center}
\includegraphics[
                width=\columnwidth,
                 ]{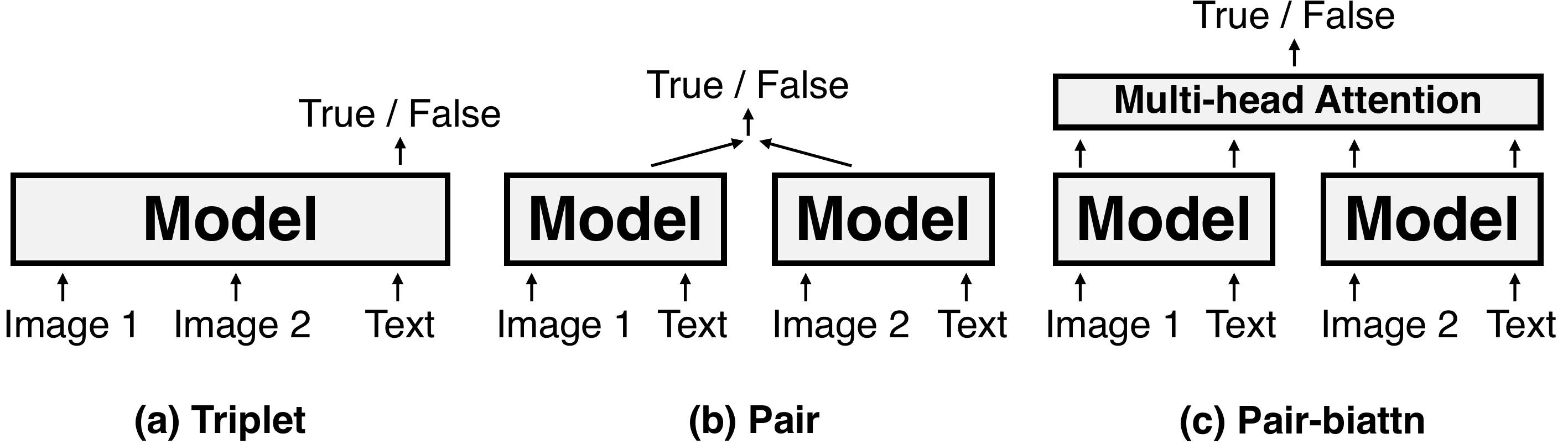}
\vspace{-15pt}
\caption{
Different encoding settings for \NLVR{}.
\textit{Pair} and \textit{Pair-biattn} approximately double the computational cost over \textit{Triplet}, which our models are based on.}
\label{fig:nlvr_settings}
\end{center}
\vspace{-15pt}
\end{figure}

%% file: table_refcoco.tex
\begin{table}[t]
\caption{RefCOCOg performance comparison.}
\label{table:refcoco}
\vskip 0.15in
\centering
\begin{tabular}{lccc}
\toprule
Method & V\&L PT & val$^d$ & test$^d$ \\
\midrule
MattNet & & 66.9 & 67.3  \\
UNITER$_{Base}$ & \checkmark & \textbf{74.3} & \textbf{74.5} \\
\midrule
\ourst{} & & 63.4 & 62.9  \\
\ourst{} & \checkmark & 71.2 & 71.3  \\
\oursb{} &  & 21.8 & 23.0  \\
\oursb{} & \checkmark & 23.6 & 22.4  \\
\bottomrule
\vspace{-20pt}
\end{tabular}
\end{table}

%% file: table_vcr_full.tex
\begin{table*}[t]
\caption{
VCR accuracy. \textit{Stage 1} refers to the original generic-domain pretraining and \textit{Stage 2} refers to the in-domain pretraining on VCR.}
\label{table:vcr_full}
\vskip 0.15in
\centering
\resizebox{.86\textwidth}{!}{
\begin{tabular}{l cc ccc ccc}
\toprule
\multirow{2}{*}{Method} &  \multicolumn{2}{c}{V\&L PT} & \multicolumn{3}{c}{VCR val} & \multicolumn{3}{c}{VCR test} \\
 & Stage 1 & Stage 2 & Q $\rightarrow$ A & QA $\rightarrow$ R & Q $\rightarrow$ AR & Q $\rightarrow$ A & QA $\rightarrow$ R & Q $\rightarrow$ AR \\
\midrule
ViLBERT & & & 69.3 & 71.0 & 49.5 & - & - & - \\
ViLBERT &\checkmark & & 72.4 & 74.4 & 54.0 & 73.3 & 74.6 & 54.8 \\
UNITER$_{Base}$ & & & 72.4 & 73.7 & 53.5 & - & - & - \\
UNITER$_{Base}$ & \checkmark & & 72.8 & 75.3 & 54.9 & - & - & - \\
UNITER$_{Base}$ & \checkmark & \checkmark & \textbf{74.6} & \textbf{77.0} & \textbf{57.8} & 75.0 & 77.2 & 58.2 \\
\midrule
\ourst{} & & & 71.1 & 73.6 & 52.5 & - & - & - \\
\ourst{} & \checkmark & & 72.9 & 75.0 & 54.7 & - & - & - \\
\ourst{} & \checkmark & \checkmark & \textbf{74.6} & \textbf{77.0} & \textbf{57.5} & \textbf{75.3} & \textbf{77.8} & \textbf{58.9} \\
\oursb{} & & & 65.4 & 68.1 & 44.6 & - & - & - \\
\oursb{} & \checkmark & & 67.0 & 67.4 & 45.4 & - & - & - \\
\oursb{} & \checkmark & \checkmark  & 69.2 & 69.9 & 48.6 & 69.8	& 69.8 & 48.9 \\
\bottomrule
\vspace{-20pt}
\end{tabular}
}
\end{table*}

%% file: table_caption.tex
\begin{table}[t]
\caption{
COCO captioning scores on Karparthy-test split. All models are trained with cross-entropy loss. PT and FT refer to the use of object tags during pretraining and finetuning, respectively.}
\label{table:caption}
\vskip 0.15in
\centering
\resizebox{\columnwidth}{!}{
\begin{tabular}{l cc cccc}
\toprule
\multirow{2}{*}{Method} & \multirow{2}{*}{V\&L PT} & \multirow{2}{*}{Object tags} & \multicolumn{4}{c}{COCO Captioning} \\
& & & B & C & M & S \\
\midrule
Oscar           & \checkmark & PT+FT        & \textbf{36.5} & \textbf{123.7} & \textbf{30.3} & \textbf{23.1} \\
\ourst{}        & \checkmark & FT           & 34.5 & 116.5 & 28.7 & 21.9 \\
\oursb{}        & \checkmark & FT           & 35.1 & 116.6 & 28.7 & 21.5 \\
\midrule
Oscar           & \checkmark &              & 34.5 & 115.6 & \textbf{29.1} & \textbf{21.9} \\ 
Unified VLP     & \checkmark &              & 36.5 & 117.7 & 28.4 & 21.3 \\
XGPT            & \checkmark &              & \textbf{37.2} & \textbf{120.1} & 28.6 & 21.8 \\
\ourst{}        & \checkmark &              & 34.6 & 116.1 & 28.8 & \textbf{21.9} \\
\oursb{}        & \checkmark &              & 34.2 & 114.1 & 28.4 & 21.3 \\
\midrule
Unified VLP     & &                         & 35.5 & \textbf{114.3} & 28.2 & 21.0 \\
XGPT            & &                         & 34.4 & 113.0 & 27.8 & 20.8 \\
BUTD            &  &                        & \textbf{36.2} & 113.5 & 27.0 & 20.3  \\
\ourst{}        & &                         & 32.6 & 109.4 & 28.2 & 21.0 \\
\oursb{}        & &                         & 33.8 & 112.4 & \textbf{28.5} & \textbf{21.4} \\
\bottomrule
\vspace{-20pt}
\end{tabular}
}
\end{table}

%% file: table_mmt.tex
\begin{table}[]
\caption{Multi30K En-De multimodal translation BLEU scores. $\dagger$ and * refer to data augmentation and ensemble, respectively.
We use gray color for the ensemble model it is not fairly comparable.
}
\label{table:mmt}
\vskip 0.15in
\centering
\resizebox{\columnwidth}{!}{
\begin{tabular}{lcccc}
\toprule
Method & V\&L PT & test2016 & test2017 & test2018 \\
\midrule
MSA & & 38.7 & - & - \\
MeMAD & & 38.9 & 32.0 & -  \\
\midrule
MSA$^\dagger$ & & 39.5 & - & -  \\
MeMAD$^{\dagger}$ & & 45.1 & 40.8 & -   \\
\demph{MeMAD$^{\dagger*}$} & & \demph{45.5} & \demph{41.8} & \demph{38.5}  \\
\midrule
T5 (text only) & & 44.6 & 41.6 & 39.0   \\
\ourst{} & & 45.3 & \textbf{42.4} & \textbf{39.5}  \\
\ourst{} & \checkmark & \textbf{45.5} & 40.9 & 38.6  \\
BART (text only) & & 41.2 & 35.4 & 33.3 \\
\oursb{} & & 41.3 & 35.9 & 33.2   \\
\oursb{} & \checkmark & 37.7 & 29.7 & 28.1  \\
\bottomrule
\vspace{-18pt}
\end{tabular}
}
\end{table}

%% file: table_multitask_all.tex
\begin{table*}[]
\caption{Single-task vs. Multi-task finetuning results on 7 tasks.
With a single set of parameters, our multi-task model achieves similar performance to separately optimized single-task models.
We denote the number of parameters of single \ourst{} model as P.
}
\label{table:multitask-all}
\vskip 0.15in
\centering
\resizebox{\textwidth}{!}{
\begin{tabular}{lccccccccc}
\toprule
\multirow{4}{*}{Method} & \multirow{4}{1.4cm}{\centering Finetuning tasks} & \multirow{4}{*}{\# Params} & \multicolumn{5}{c}{Discriminative tasks} & \multicolumn{2}{c}{Generative tasks} \\
\cmidrule(lr){4-8} \cmidrule(lr){9-10}
 & & & VQA & GQA & \NLVR{} & RefCOCOg & VCR & COCO Caption & Multi30K En-De \\
 & & & Karpathy test & test-dev & test-P & test$^d$ & val & Karpathy test & test2018 \\
 & & & Acc & Acc & Acc & Acc & Acc & CIDEr & BLEU\\
\midrule
\ourst{} & single task & 7P & 67.9 & 60.0 & 73.6 & 71.3 & 57.5 & 116.1 & 38.6 \\
\ourst{} & all tasks & P & 67.2 & 58.9 & 71.6 & 69.4 & 55.3 & 110.8 & 37.6 \\
\bottomrule
\vspace{-20pt}
\end{tabular}
}
\end{table*}

%% file: table_multitask_head.tex
\begin{table}[t]
\caption{Multi-task finetuning with single/task-specific heads. While three tasks are included for brevity, the rest of the tasks also show the minimal differences between two setups.}
\label{table:multitask-FT-small}
\vskip 0.1in
\centering
\resizebox{\columnwidth}{!}{
\begin{tabular}{ccccc}
\toprule
\multirow{3}{*}{Method} & \multirow{3}{*}{\# Params} & VQA & GQA & COCO Caption  \\
 & & Karpathy test & test-dev & Karpathy test  \\
 & & Acc & Acc & CIDEr \\
\midrule
Single shared head & P & 68.3 & 59.3 & 110.6 \\
Task-specific heads & P+7H=1.8P & 68.5 & 59.3 & 110.9 \\
\bottomrule
\end{tabular}
}
\vspace{-15pt}
\end{table}

%% file: conclusion.tex
\section{Conclusion}
In this work, we proposed \ourst{} and \oursb{} which tackle vision-and-language tasks with a unified text generation objective.
Experiments show \ourst{} and \oursb{} can achieve comparable performance with state-of-the-art vision-and-language transformers on diverse vision-and-language tasks without hand-crafted architectures and objectives.
Especially, we demonstrate our generative approach is better suited for open-ended visual question answering.
In addition, we also showed it is possible to train seven different tasks simultaneously using a single architecture with single parameters without not losing much performance. It would be an interesting future work to further explore this direction by adding even more tasks.

\section*{Acknowledgments}
We thank Hyounghun Kim, Zineng Tang, Swarnadeep Saha, Xiang Zhou, and anonymous reviewers for their comments and suggestions. This work was supported by NSF-CAREER Award 1846185, ARO-YIP Award W911NF-18-1-0336, DARPA MCS Grant N66001-19-2-4031, Google Focused Research Award, and Bloomberg Data Science Ph.D. Fellowship. The views, opinions, and/or findings contained in this article are those of the authors and not of the funding agency.

%% file: supplementary.tex
\section{Comparison with Baselines}

In Table \ref{table:baselines}, we compare the baseline vision-and-language transformers with our \ourst{} and \oursb{} in detail, including their pretraining datasets, architecture, etc. 

\input{table_baselines}

\section{Implementation Details}

In Table~\ref{table:task_pretraining} and Table~\ref{table:dataset_downstream}, we show the detailed statistics of our pretraining and downstream datasets and tasks.
In Table~\ref{table:hyperparameter}, we show the hyperparameters that we used in our pretraining and downstream task experiments. We provide the links to download pretraining and downstream datasets.

\input{table_data_pretraining}

\input{table_data_downstream}
\input{table_hyperparameter}

\subsection{Pretraining Data}
Overall, our pretraining dataset contains 9.18M image-text pairs on 180K distinct images.
We carefully split our pretraining data to avoid any intersection between our training data and the validation/test sets of the downstream tasks (e.g., COCO Captioning, RefCOCOg). 
In this process, around 10K images are excluded from the training sets of COCO\footnote{\url{https://cocodataset.org/\#download}} and Visual Genome\footnote{\url{http://visualgenome.org/api/v0/api_home.html}}. 
We use COCO \textit{Karpathy val split} \cite{Karpathy2015} with 5,000 images as our validation set to monitor pretraining performance.

\subsection{Downstream Tasks}

\paragraph{VQA\footnote{\url{https://visualqa.org/download.html}}, COCO caption}
For both VQA and COCO captioning tasks, we follow \textit{Karparthy split} \cite{Karpathy2015}, which re-splits train2014 and val2014 COCO images \cite{Lin2014} into 113,287 / 5,000 / 5,000 images for train / validation / test.

\paragraph{GQA\footnote{\url{https://cs.stanford.edu/people/dorarad/gqa/download.html}}}
Following LXMERT \cite{Tan2019}, we use GQA-balanced version.
We use train and val splits for training and use test-dev split for validation.
Train / val / test-dev splits consist of 943,000 / 132,062 / 12,578 questions, respectively.

\paragraph{\NLVR{} \footnote{\url{http://lil.nlp.cornell.edu/nlvr/}}}
Train / val / test-P splits consist of 86,373 / 6982 / 6967 sentences, respectively.
We train our model on train split and use val split for validation.

\paragraph{VCR\footnote{\url{https://visualcommonsense.com/download/}}}
Train / val / test splits consist of 212,923 / 26,534 / 25,263 questions, respectively.
We train our model on train split and use val split for validation.

\paragraph{RefCOCOg\footnote{\url{https://github.com/lichengunc/refer}}}
We use \textit{umd} split, which consists of train / val / test sets with 42,226 / 2,573 / 5,023 sentences, respectively.
Following UNITER \cite{Chen2020} and MAttNet \cite{Yu2018}, 
we use ground truth COCO boxes for training,
and use the detected boxes from an off-the-shelf Mask R-CNN \footnote{\url{https://github.com/lichengunc/MAttNet\#pre-computed-detectionsmasks}} as candidates during inference.

\paragraph{Multi30K En-De\footnote{\url{https://github.com/multi30k/dataset}}}
The train / val / test2016 / test2017 / test2018 splits consist of 29,000 / 1,014 / 1,000 / 1,000 / 1,017 English-German sentence pairs, respectively.

%% file: table_baselines.tex
\begin{table*}[!thbp]
\caption{Summary of baseline vision-and-language transformers.
$^a$Since not all models report exact parameter numbers, we provide rough estimates compared to BERT$_{Base}$ (86M; noted as P), where word embedding parameters are excluded.
$^b$LXMERT and XGPT are not initialized from pretrained language models.
LXMERT authors found pretraining from scratch was more effective than initialization from BERT$_{Base}$ in their experiments.
XGPT uses text pretraining on Conceptual captions and COCO captions with Masked LM \cite{Devlin2019} and Masked Seq2Seq \cite{Song2019} objectives before V\&L pretraining.
$^c$LXMERT (text+visual+cross-modal) and ViLBERT (cross-modal) use dual-stream encoders.
ViLBERT uses 768/1024-dim hidden states for text/visual streams respectively.
XGPT uses AoA module \cite{Huang2019} as visual encoder.
Rest of the models use single-stream encoders.
$^d$For generation tasks, Unified VLP and Oscar use causal mask and reuse encoder as decoder similar to UniLM.
$^e$XGPT also uses shared parameters for encoder and decoder, but its decoder is right-shifted and predicts next tokens.
$^f$Unified VLP is initialized from UniLM, which is initialized from BERT$_{Large}$.
$^g$Oscar uses object tags as additional text inputs.
}
\label{table:baselines}
\vskip 0.15in
\centering
\resizebox{\textwidth}{!}{
\begin{tabular}{llc llccccc}
\toprule
 & \multicolumn{2}{c}{V\&L Pretraining} & \multicolumn{7}{c}{Hyperparameters} \\
 \cmidrule(lr){2-3} \cmidrule(lr){4-10}
 & Dataset & \# Imgs & Arch. type & Backbone & \# Layers & \# Params$^a$ & Hidden dim & \# Regions & Position Emb \\
\midrule
\midrule
LXMERT  & COCO+VG & 180K & Encoder & -$^b$ & 9+5+5$^c$ & 2P & 768 & 36 & absolute\\
ViLBERT & CC & 3M & Encoder & BERT$_{Base}$ & 12$^c$ & 2.5P & 768/1024$^c$ & 10$\sim$36 & absolute \\
UNITER$_{Base}$  & CC+SBU+COCO+VG & 4M & Encoder & BERT$_{Base}$ & 12 & P & 768 & 10$\sim$100 & absolute \\
Unified VLP & CC & 3M & Encoder$^d$ & UniLM$^f$ & 12 & P & 768 & 100 & absolute \\
Oscar$_{Base}$ & CC+SBU+COCO+VG+Flickr30K & 4M & Encoder$^d$ & BERT$_{Base}$ & 12 & P & 768 & 50$^g$ & absolute  \\
XGPT & CC+COCO & 3M & Enc-Dec$^e$ & -$^b$ & 1$^c$+12+12 & P & 768 & 100 & absolute \\
\midrule
\ourst{}  & COCO+VG & 180K & Enc-Dec & T5$_{Base}$ & 12+12 & 2P & 768 & 36 & relative \\
\oursb{}  & COCO+VG & 180K & Enc-Dec & BART$_{Base}$ & 6+6 & P & 768 & 36 & absolute \\
\bottomrule
\end{tabular}
}
\end{table*}

%% file: table_data_pretraining.tex
\begin{table*}[!htbp]
\caption{
Pretraining tasks used in our vision-and-language pretraining.
The images that have any intersection with evaluation set of downstream tasks (e.g., COCO caption, RefCOCOg) and the held-out validation set for pretraining are excluded.
}
\label{table:task_pretraining}
\vskip 0.15in
\centering
\begin{tabular}{lllccl}
\toprule
Task & Image source & Text source & \# Examples  \\
\midrule
Multimodal language modeling & COCO, VG & COCO caption, VG caption & 4.9M (\# captions) \\
Visual question answering & COCO, VG & VQA, GQA, Visual7W & 2.5M (\# questions) \\
Image-text matching & COCO & COCO caption & 533K (\# captions) \\
Visual grounding & COCO, VG & object\&attribute tags & 163K (\# images) \\
Grounded captioning & COCO, VG & object\&attribute tags & 163K (\# images) \\
\bottomrule
\end{tabular}
\end{table*}

%% file: table_data_downstream.tex
\begin{table*}[!htbp]
\caption{Statistics of the datasets used in downstream tasks.
The data that are not used for training/validation (e.g., COCO test2015 images) and data for leaderboard submissions (e.g., test-dev/test-std for VQA, test for GQA) are excluded.
}
\label{table:dataset_downstream}
\vskip 0.15in
\centering
\begin{tabular}{llccl}
\toprule
Datasets & Image source & \# Images (train) & \# Text (train) & Metric \\
\midrule
VQA & COCO & 123K (113K) & 658K (605K) & VQA-score \\
GQA & VG & 82.7K (82.3K) & 1.08M (1.07M) & Accuracy \\
\NLVR{} & Web Crawled & 238K (206K) & 100K (86K) & Accuracy \\
RefCOCOg & COCO & 26K (21K) & 95K (80K) & Accuracy \\
VCR & Movie Clips & 110K (80K) & 290K (212K) & Accuracy \\
COCO Caption & COCO & 123K (113K) & 616K (566K) & BLEU,CIDEr,METEOR,SPICE \\
Multi30K En-De & Flickr30K & 31K (29K) & 31K (29K)  & BLEU \\
\bottomrule
\end{tabular}
\end{table*}

%% file: table_hyperparameter.tex
\begin{table*}[!htbp]
\caption{Hyperparameters for pretraining and downtream tasks}
\label{table:hyperparameter}
\vskip 0.15in
\centering
\begin{tabular}{ll ccc}
\toprule
Model & Task & Learning rate & Batch size & Epochs \\
\midrule
\multirow{9}{*}{\ourst{}} & Pretraining & 1e-4 & 320 & 30 \\
& VCR Pretraining & 5e-5 & 80 & 20 \\
& VQA & 5e-5 & 320 & 20 \\
& GQA &  1e-5 & 240 & 20 \\
& \NLVR{} & 5e-5 & 120 & 20 \\
& RefCOCOg & 5e-5 & 360 & 20 \\
& VCR &  5e-5 & 16 & 20 \\
& COCO Caption & 3e-5 & 320 & 20 \\
& Multi30K En-De &  5e-5 & 120 & 20 \\
\midrule
\multirow{9}{*}{\oursb{}} & Pretraining & 1e-4 & 600 & 30 \\
& VCR Pretraining & 5e-5 & 120 & 20 \\
& VQA & 5e-5 & 600 & 20 \\
& GQA &  1e-5 & 800 & 20 \\
& \NLVR{} & 5e-5 & 400 & 20 \\
& RefCOCOg & 5e-5 & 1200 & 20 \\
& VCR &  5e-5 & 48 & 20 \\
& COCO Caption & 3e-5 & 520 & 20 \\
& Multi30K En-De & 5e-5 & 320 & 20 \\
\bottomrule
\end{tabular}
\end{table*}